\documentclass[11pt]{article}
\usepackage{coling2020}
\usepackage{times}
\usepackage{url}
\usepackage{latexsym}

\usepackage{graphicx}
\usepackage{makecell}
\usepackage{booktabs}
\usepackage{amsmath}
\usepackage{amssymb}
\usepackage{multirow}
\usepackage[dvipsnames]{xcolor}
\usepackage[colorlinks]{hyperref}
\hypersetup{
linkcolor=NavyBlue
,citecolor=NavyBlue
,filecolor=Red
,urlcolor=NavyBlue
,menucolor=Red
,runcolor=Red
,linkbordercolor=NavyBlue
,citebordercolor=NavyBlue
,filebordercolor=Red
,urlbordercolor=NavyBlue
,menubordercolor=Red
,runbordercolor=Red
}
\usepackage{subcaption}
\usepackage{pifont}

\colingfinalcopy 

\title{Incorporating Commonsense Knowledge into Abstractive Dialogue Summarization via Heterogeneous Graph Networks}

\author{Xiachong Feng, Xiaocheng Feng, Bing Qin, Ting Liu \\
Research Center for Social Computing and Information Retrieval, \\
Harbin Institute of Technology, China \\
 
 {\tt \{xiachongfeng,xcfeng,qinb,tliu\}@ir.hit.edu.cn} \\
}

\date{}

\begin{document}
\maketitle
\begin{abstract}
Abstractive dialogue summarization is the task of capturing the highlights of a dialogue and rewriting them into a concise version.
In this paper, we present a novel multi-speaker dialogue summarizer to demonstrate how large-scale commonsense knowledge can facilitate dialogue understanding and summary generation.
In detail, we consider utterance and commonsense knowledge as two different types of data and design a Dialogue Heterogeneous Graph Network (D-HGN) for modeling both information. Meanwhile, we also add speakers as heterogeneous nodes to facilitate information flow. Experimental results on the SAMSum dataset show that our model can outperform various methods. We also conduct zero-shot setting experiments on the Argumentative Dialogue Summary Corpus, the results show that our model can better generalized to the new domain.
\end{abstract}

\begin{figure}[b]
	\centering
	\includegraphics[scale=0.41]{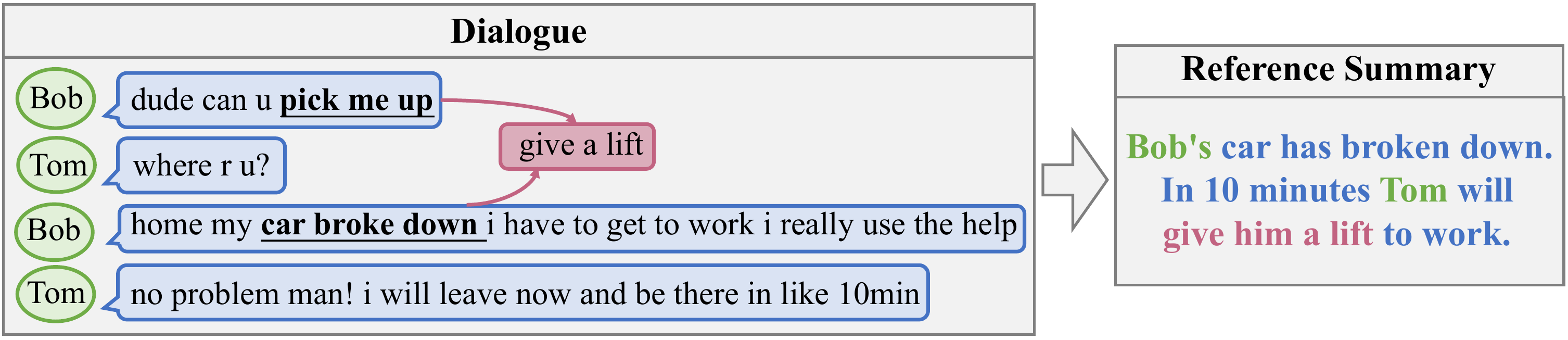}
	\caption{An example of dialogue-summary pair. Green for speakers, blue for utterances, and pink for commonsense knowledge. In order to generate ``give a lift" in the reference summary, the summarization model needs to understand the commonsense knowledge behind ``pick up" and ``car broke down".}
	\label{fig:intro}
\end{figure}

\section{Introduction}
Automatic summarization is a fundamental task in Natural Language Processing, which aims to condense the original input into a shorter version covering salient information and has been continuously studied for decades \cite{old2,old3}. Recently, online multi-speaker dialogue/meeting has become one of the most important ways for people to communicate with each other in their daily works. Especially due to the spread of  COVID-19 worldwide, people are more dependent on online communication. In this paper, we focus on dialogue summarization, which can help people quickly grasp the core content of the dialogue without reviewing the complex dialogue context.

Recent works that incorporate additional commonsense knowledge in the dialogue generation \cite{knowledge-dg} and dialogue context representation learning \cite{aaai20-pretrain} show that even though neural models have strong learning capabilities, explicit knowledge can still improve response generation quality.  
It is because that a dialog system can understand conversations better and thus respond more properly if it can access and make full use of large-scale commonsense knowledge.
However, current dialogue summarization systems \cite{crf-discourse,acl19li,kdd19,hmnet} ignore the exploration of commonsense knowledge, which may limit the performance.
In this work, we examine the benefit of incorporating commonsense knowledge in the dialogue summarization task and also address the question of how best to incorporate this information.
Figure \ref{fig:intro} shows a positive example to illustrate the effectiveness of commonsense knowledge in the dialogue summarization task. 
Bob asks Tom for help because his car has broken down. On the one hand, by introducing commonsense knowledge according to the {\em pick up} and {\em car broke down}, we can know that Bob expects Tom to {\em give him a lift}. On the other hand, commonsense knowledge can serve as a bridge between non-adjacent utterances that can help the model better understanding the dialogue.

In this paper, we follow the previous setting \cite{knowledge-dg} and also use ConceptNet \cite{conceptnet} as a large-scale commonsense knowledge base, while the difference is that we regard knowledge and text(utterance) as heterogeneous data in a real multi-speaker dialogue. We propose a model named \textbf{D}ialogue \textbf{H}eterogeneous \textbf{G}raph \textbf{N}etwork (D-HGN) for incorporating commonsense knowledge by constructing the graph including both utterance and knowledge nodes. Besides, our heterogeneous graph also contains speaker nodes at the same time, which has been proved to be a useful feature in dialogue modeling. In particular, we equip our heterogeneous graph network with two additional designed modules. One is called message fusion, which is specially designed for utterance nodes to better aggregate information from both speakers and knowledge. The other one is called node embedding, which can help utterance nodes to be aware of position information. Compared to homogeneous graph network in related works \cite{crf-discourse,acl19li,kdd19,hmnet}, we claim that the heterogeneous graph network can effectively fuse information and contain rich semantics in nodes and links, and thus more accurately encode the dialogue representation.

We conduct experiments on the SAMSum corpus \cite{samsum}, which is a large-scale chat summarization corpus. We analyze the effectiveness of integration of knowledge and heterogeneity modeling. The human evaluation also shows that our approach can generate more abstractive and correct summaries. To evaluate whether commonsense knowledge can help our model better generalize to the new domain, we also perform zero-shot setting experiments on the Argumentative Dialogue Summary Corpus \cite{adsc}, which is a debate summarization corpus. In the end, we give a brief summary of our contributions: (1) We are the first to incorporate commonsense knowledge into dialogue summarization task. (2) We propose a D-HGN model to encode the dialogue by viewing utterances, knowledge and speakers as heterogeneous data. (3) Our model can outperform various methods.

\section{Heterogeneous Dialogue Graph Construction}
In this section, we describe the graph notation and the graph construction process, which consists of three steps, including (1) utterance-knowledge bipartite graph construction, (2) speaker-utterance bipartite graph construction and (3) heterogeneous dialogue graph construction. 

\subsection{Graph Notation}
Our heterogeneous dialogue graph (HDG) is defined as a directed graph $G=(\mathcal{V}, \mathcal{E}, \mathcal{A}, \mathcal{R})$, where each node $v \in \mathcal{V}$ and each edge $e \in \mathcal{E}$. Different types of nodes and edges are associated with their type mapping functions $\tau(v): \mathcal{V} \rightarrow \mathcal{A}$ and $\phi(e): \mathcal{E} \rightarrow \mathcal{R}$.

\subsection{Utterance-Knowledge Bipartite Graph Construction}
Current dialogue summarization corpus has no knowledge annotations. To ground each dialogue to commonsense knowledge, we make use of ConceptNet \cite{conceptnet} to incorporate knowledge. ConceptNet is a semantic network that contains 34 relations in total and represents each knowledge tuple by $R=(h, r, t, w)$ meaning that head concept $h$ and tail concept $t$ have a relation $r$ with a weight of $w$. It contains not only world facts such as ``\textit{Paris is the capital of France}" that are constantly true, but also informal relations that are part of daily knowledge such as ``\textit{Call is used for Contact}". 

\begin{figure}[t]
	\centering
	\includegraphics[scale=0.42]{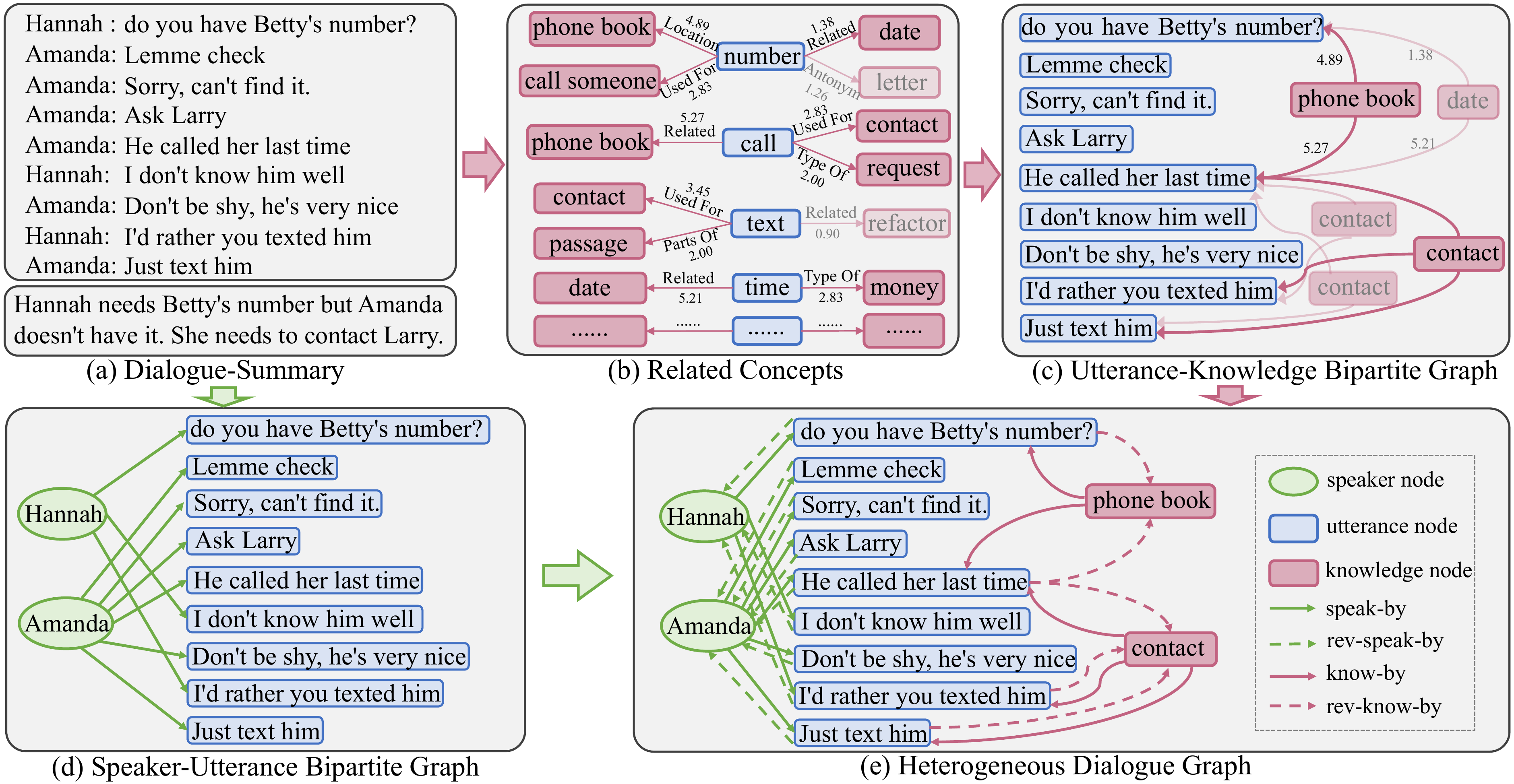}
	\caption{Illustration of Heterogeneous Dialogue Graph Construction Process.}
	\label{fig:graph_construction}
\end{figure}

We use each word in the utterance as a query to retrieve a one-hop graph from ConceptNet, as done by \newcite{StoryEG}. We only consider nouns, verbs, adjectives, and adverbs. We filter out tuples where (1) $r$ is in a pre-defined list of useless relations\footnote{We pre-define the useless relation list, including Antonym, EtymologicallyDerivedFrom, NotHasProperty, DistinctFrom, NotCapableOf, EtymologicallyRelatedTo and NotDesires.} (e.g. ``number" is antonym of ``letter"), (2) the weight of $r$ is less than 1 (e.g. ``text" is related to ``refactor", weight: 0.9). Finally, we can get related concepts for the dialogue, as shown in Figure \ref{fig:graph_construction}(b). We construct utterance-knowledge bipartite graph by viewing utterances and knowledge as different types of nodes. As shown in Figure \ref{fig:graph_construction}(c), we connect two utterances to one tail concept $t$ using edge {\em know-by} if they both have the same tail concept $t$. Note that two utterances may connect to multiple tail concepts, we choose the one with the highest average weight of relations (e.g. ``phone book" is better than ``date"). If there are multiple identical knowledge nodes, we also combine them to a single one (e.g. two ``contact" nodes are combined into one node).

\subsection{Speaker-Utterance Bipartite Graph Construction}
Given multiple speakers and corresponding utterances in a dialogue, we construct the speaker-utterance bipartite graph by viewing speakers and utterances as different types of nodes. As shown in Figure \ref{fig:graph_construction}(d), we construct {\em speak-by} edges from speakers to utterances based on who said the utterances.

\subsection{Heterogeneous Dialogue Graph Construction}
We combine the utterance-knowledge bipartite graph and the speaker-utterance bipartite graph as our heterogeneous dialogue graph, as shown in Figure \ref{fig:graph_construction}(e). Additionally, we add a reverse edge {\em rev-know-by} and {\em rev-speak-by} to facilitate information flow over the graph. Finally, there are three types of nodes, where $\mathcal{A}$ becomes {\em speaker}, {\em utterance}, and {\em knowledge} and four types of edges, where  $\mathcal{R}$ becomes {\em speak-by}, {\em know-by}, {\em rev-speak-by} and {\em rev-know-by}. 

\section{Dialogue Heterogeneous Graph Network}
\begin{figure}[t]
	\centering
	\includegraphics[scale=0.68]{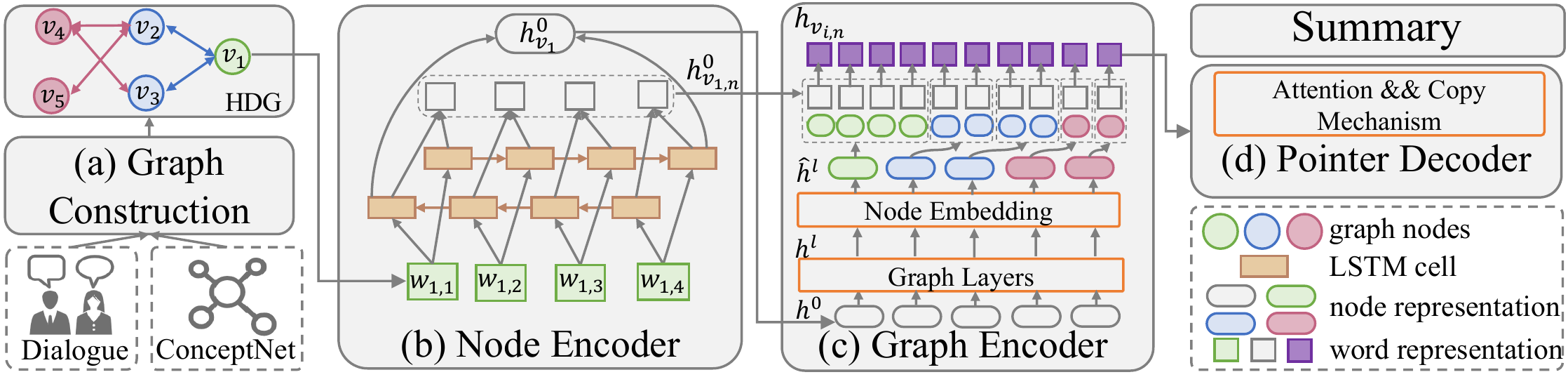}
	\caption{Illustration of our D-HGN model. (a) Graph construction receives a dialogue and ConceptNet and outputs a heterogeneous dialogue graph (HDG). (b) Node encoder receives a sequence of words for a node and produces initial node and word representations. (c) Graph encoder first conducts graph operations for initial node representations. Then a node embedding module is added after graph layers to make nodes to be aware of position information. Finally, the initial word representations and corresponding updated node representations are concatenated as final word representations. (d) Pointer decoder can either generate summary words from the vocabulary or copy from the input words.}
	\label{fig:framework}
\end{figure}

In this section, we describe the details of our dialogue heterogeneous graph network (D-HGN), including three components: node encoder, graph encoder and pointer decoder. The model is shown in Figure \ref{fig:framework}.

\subsection{Node Encoder}

The role of node encoder is to give each node $v_i \in \mathcal{V}$ an initial representation $h^{0}_{v_i}$, where $v_i$ consists of $|v_i|$ words $[w_{i,1},w_{i,2},...w_{i,|v_i|}]$. Note that speaker and knowledge may have multiple words. We employ a Bi-LSTM as the node encoder that encodes input node forwardly and backwardly to generate two sequences of hidden states $\left(\overrightarrow{h_1}, \overrightarrow{h_2}, \ldots, \overrightarrow{h_{|v_i|}}\right)$ and $\left(\overleftarrow{h_1}, \overleftarrow{h_2}, \ldots, \overleftarrow{h_{|v_i|}}\right)$, where $\overrightarrow{h_n}\!=\!\operatorname{LSTM}_{f}\left(x_{n}, \overrightarrow{h_{n-1}}\right)$ and $\overleftarrow{h_n}\!=\!\operatorname{LSTM}_{b}\left(x_{n}, \overleftarrow{h_{n+1}}\right)$. $x_n$ denotes the embedding of $w_{i,n}$. The forward and backward hidden states are concatenated as the initial node representation $h^{0}_{v_i}=[\overrightarrow{h_{|v_i|}};\overleftarrow{h_1}]$ and initial word representation $h^{0}_{v_i,n}=[\overrightarrow{h_{n}};\overleftarrow{h_n}]$. $h^{0}_{v_i}$ will be passed to the graph encoder to learn high-level representations. $h^{0}_{v_i,n}$ will be concatenated with updated node representations to get final word representations. 

\subsection{Graph Encoder}
Graph encoder is used to digest the structural information and get updated node representations. We employ Heterogeneous Graph Transformer \cite{hgt} as our graph encoder, which  models heterogeneity by type-dependent parameters and can be easily applied to our graph. It includes: (a) heterogeneous mutual attention, which calculates attention scores $\operatorname{Attn}(s, e, t)$ between source nodes and the target node. (b) heterogeneous message passing, which prepares the message vector $\operatorname{Msg}(s, e, t)$ for each source node and  (c) target-specific aggregation, which aggregates messages from source nodes to the target node 
using attention scores as the weight. Specifically, we design two modules named message fusion and node embedding to make the learning process more effective for our graph.

\begin{figure*}[t]
	\centering
	\includegraphics[scale=0.86]{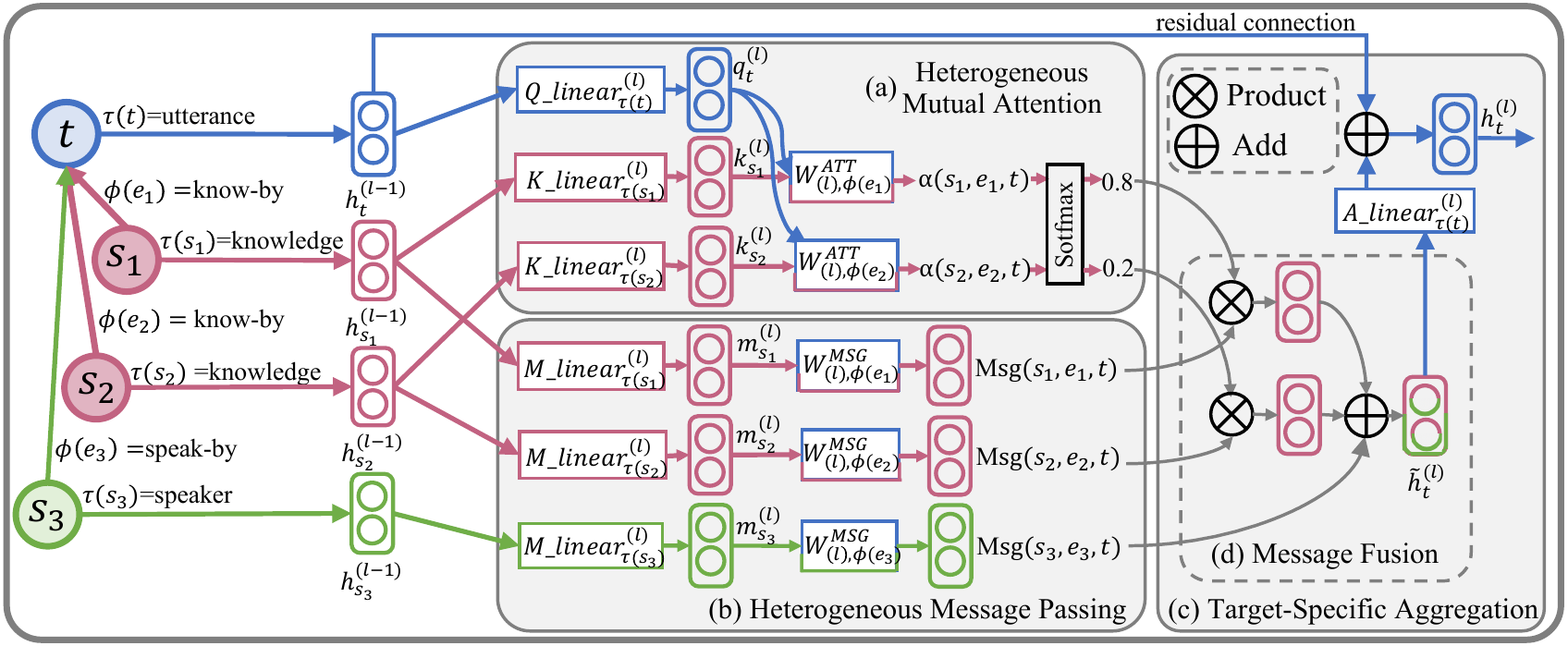}
	\caption{Illustration of one graph layer. Given a target node of utterance type and source nodes of knowledge and speaker type. Firstly, we use (a) heterogeneous mutual attention to calculate the attention scores by type-dependent linear projection. Secondly, we use (b) heterogeneous message passing to prepare the message vector for each source node. Thirdly, we use (c) target-specific aggregation to aggregate messages to the target node. Specifically, we propose a message fusion module that uses attention scores as the weight to average the knowledge vectors and add speaker information additionally. }
	\label{fig:hgt}
\end{figure*}

\noindent \textbf{Heterogeneous Mutual Attention} Given an edge $e=(s,t)$ with their node and edge type mapping functions $\tau$ and $\phi$, we first project source and target node representations from $(l$-$1)$-th layer $h^{(l-1)}_{s}$ and $h^{(l-1)}_{t}$ into key vector $k_s^{(l)} =\mathrm{K}\_\operatorname{Linear}_{\tau(s)}^{(l)}\left(h^{(l-1)}_{s}\right)$ and query vector $q_t^{(l)} =\mathrm{Q}\_\operatorname{Linear}_{\tau(t)}^{(l)}\left(h^{(l-1)}_{t}\right)$ with 
type-dependent linear projection. Next, to integrate edge type information, we calculate unnormalized score $\alpha(s, e, t)$ between $t$ and $s$ by adding a edge-based matrix $W_{(l),\phi(e)}^{ATT}$. Finally, for each target node $t$, we conduct $\operatorname{Softmax}$ for all $s \in N(t)$ to get the final normalized attention scores $\operatorname{Attn}^{(l)}(s, e, t)$, where $N(t)$ denotes neighbors of target node $t$. Note that if target node is of utterance type and source node is of speaker type, we do not calculate the attention score between these two types of nodes. See more detail at {\em message fusion} module. The process is shown in Figure \ref{fig:hgt}(a).
\begin{equation}
\begin{split}
& \alpha(s,e,t) =\left(k_s^{(l)} W_{(l),\phi(e)}^{ATT} {q_t^{(l)}}^{\top}\right) \\
& \operatorname{Attn}^{(l)}(s, e, t)=\mathop{\operatorname{Softmax}}\limits_{\forall s \in N(t)}\left(\alpha(s, e, t)\right) \\
\end{split}
\end{equation}

\noindent \textbf{Heterogeneous Message Passing} We first project source node representation $h^{(l-1)}_{s}$ into the vector $m_s^{(l)}=\mathrm{M}\_\operatorname{Linear}_{\tau(s)}^{(l)}\left(h^{(l-1)}_{s}\right)$ with type-dependent linear projection and then followed by a edge-based matrix $W_{(l),\phi(e)}^{MSG}$ to get the message vector. The process is shown in Figure \ref{fig:hgt}(b).
\begin{equation}
\begin{split}
& \operatorname{Msg}^{(l)}(s, e, t)=m_s^{(l)}W_{(l),\phi(e)}^{MSG}
\end{split}
\end{equation}

\noindent \textbf{Target-Specific Aggregation}  We divide this process into two cases based on the type of target node: (1) $\tau(t) \!\neq\! utterance$, (2) $\tau(t)\!=\!utterance$. For the first case, We use attention vector as the weight to average messages: $\widetilde{h}^{(l)}_t\!=\!\mathop{\oplus}_{\forall s \in N(t)}\left(\operatorname{Attn}^{(l)}(s, e, t) \otimes \operatorname{Msg}^{(l)}(s, e, t)\right)$. For the second case, we design a Message Fusion module to aggregate messages to utterance node more effectively. 
After getting aggregated message vector $\widetilde{h}^{(l)}_t$, we maps it back to $\tau(t)$-type distribution with a linear projection followed by residual connection: $ h^{(l)}_t=\mathrm{A}\_\operatorname{Linear}_{\tau(t)}^{(l)}\left(\operatorname{Sigmoid}\left(\widetilde{h}^{(l)}_t\right)\right)+h^{(l-1)}_t$, as shown in Figure \ref{fig:hgt}(c).

\noindent \textbf{Message Fusion} 
Dialogue summaries often describe ``{\em who did what}'', thus speaker information is required for utterances. However, if target node of utterance type aggregates messages from source nodes of knowledge and speaker type, it will prefer more to the speaker node while giving up using knowledge nodes, since attention is a normalized distribution. Therefore, in our message fusion module, we use attention weights for knowledge nodes to average corresponding messages and add 
speaker information additionally. The process is shown in Figure \ref{fig:hgt}(d).
\begin{equation}
\begin{split}
& \operatorname{s_k} = (\forall s \in N(t) \wedge \tau(s)=knowledge) , \operatorname{s_s} = (\forall s \in N(t) \wedge \tau(s)=speaker) \\
& \widetilde{h}^{(l)}_t =\mathop{\oplus}_{s \in \operatorname{s_k}}\!\left(\operatorname{Attn}^{(l)}(s, e, t) \otimes \operatorname{Msg}^{(l)}(s, e, t)\right) + \operatorname{Msg}^{(l)}(\operatorname{s_s}, e, t)
\end{split}
\label{eq:mf}
\end{equation}

\noindent \textbf{Node Embedding}
In this section,  a module named Node Embedding is designed to make utterance nodes to be aware of position information in source dialogue.
This is because original heterogeneous graph cannot directly model the chronological order between utterances, while an ideal dialogue summary needs to refer to the order of corresponding dialogue utterances.
In detail, for speaker and knowledge nodes, we fix their position to 0. For each utterance node $v_i$, it associates with a position $p_{v_i}$, which is the ranking of utterances in the original dialogue. As shown in Figure \ref{fig:framework}(c), we add position information for each node: $\hat h^{(l)}_{v_i}  = h^{(l)}_{v_i}+ W^{pos}[p_{v_i}]$, where $W^{pos}$ denotes a learnable node embedding matrix.

After getting the output representation $\hat h^{(l)}$ for each node, we concatenate updated node representation $\hat h^{(l)}_{v_i}$ and corresponding initial word representations $h^0_{v_i,n}$ followed by a linear projection $\mathrm{F}\_\operatorname{Linear}$ to get final word representations: $h_{v_i,n}=\mathrm{F}\_\operatorname{Linear}([ \hat h^{(l)}_{v_i};h^0_{v_i,n}])$.

\subsection{Pointer Decoder}
We employ a LSTM with attention and copy mechanism to generate summaries. 
At each decoding time step $t$, the LSTM reads the previous word embedding $x_{t-1}$ and previous context vector $c_{t-1}$ as inputs to compute the new hidden state $s_{t} =\operatorname{LSTM}\left(x_{t-1}, c_{t-1}, s_{t-1}\right)$. We use the average of all word representations ${s_{0}}={\operatorname{Average}}({{\sum}_{ v_i \in G}} {{\sum}_{ n \in {\left[1,|v_i|\right]}}} {h_{v_i,n}})$ in the graph to initialize the decoder. The context vector $c_t$ is computed as in \newcite{attention}, which is then used to calculate generation probability $p_{gen}$ and the final probability distribution $P(w)$, as done by \newcite{pgn}.

\subsection{Training}
For each heterogeneous dialogue graph $G$ that is paired with a ground truth summary $Y^{*}=[y^{*}_1,y^{*}_2,...,y^{*}_{|Y^{*}|}]$, we minimize the negative log-likelihood of the target words sequence.
\begin{equation}
L=-\sum_{t=1}^{|Y^{*}|} \log p\left(y_{t}^{*} | y_{1}^{*} \ldots y_{t-1}^{*}, G\right)
\end{equation}

\section{Experiments}
\noindent \textbf{Dataset} 
Following the latest works \cite{samsum,crf-discourse}, we conduct experiments on two different settings. Firstly, we train and evaluate our model on the SAMSum corpus \cite{samsum}, which contains dialogues around chit-chats topics. Secondly, we train using SAMSum corpus and use the Argumentative Dialogue Summary Corpus (ADSC) \cite{adsc} as the test set to perform zero-shot setting experiments. Each dialogue in ADSC dataset owns 5 different summaries and is mainly around debate topics. Table \ref{tab:graph_sta} shows the knowledge related statistics of two datasets.

\begin{table}[htb]
\centering
\begin{tabular}{clccc}
    \toprule
     \textbf{Dataset} &\textbf{Split} & \textbf{\#} & \textbf{Coverage} & \textbf{Average Know}  \\
    \midrule
    \multirow{3}{*}{SAMSum} & Train &14732 &94.43\% &19.60  \\
    & Valid &818 &95.72\% &18.23  \\
    & Test &819 &93.89\% &19.77  \\
    \midrule
    ADSC & Full &45 &100\% &6.50 \\
    \bottomrule 
\end{tabular}
\caption{Knowledge related statistics on SAMSum and ADSC datasets. \# is the number of dialogues. Coverage represents the percentage of dialogues with at least one knowledge node. Average Know represents the average number of knowledge nodes per dialogue.} \label{tab:graph_sta}
\end{table}

\noindent \textbf{Implementation Details}
The word embedding size is set to 100. The dimension of node encoder and pointer decoder is set to 300. The dimension of graph encoder is set to 200. The graph layer number is set to 1. Dropout is set to 0.5. We use Adam \cite{adam} with the learning rate of 0.001 and use gradient clipping with a maximum gradient norm of 2. In the test process, beam size is set to 10.

\noindent \textbf{Evaluation Metrics}
We employ the standard $F_1$ scores for ROUGE-1, ROUGE-2, and ROUGE-L metrics \cite{rouge} to measure summary qualities. These three metrics evaluate the accuracy on unigrams, bigrams, and longest common subsequence between the groundtruth and the generated summary. 

\noindent \textbf{Baseline Models}
We compare our model with several baselines. \textbf{LONGEST-3} chooses the longest three utterances as the summary. \textbf{TextRank} \cite{textrank} is a graph-based extractive method. \textbf{SummaRunner} \cite{SummaRuNNer} extract utterances based on a hierarchical RNN model. \textbf{Transformer} \cite{transformer} is a Seq2Seq model that utilizes self-attention operations. \textbf{PGN} \cite{pgn} is a Seq2Seq model equipped with copy mechanism. \textbf{HRED} \cite{HRED} is a hierarchical Seq2Seq model. \textbf{Abs RL} \cite{fastrl} is a pipeline model that first selects salient utterances 
based on a extractive model then produces the summary based on a abstractive model using diversity beam search. The extractive model is trained using utterance-level extraction labels. 
The overall model is jointly trained using reinforcement learning. Based on Abs RL, \textbf{Abs RL Enhance} \cite{samsum} appends all speakers after each utterance, because the original model may select utterances of a single speaker that will lead to no other speaker information. \textbf{D-GAT}, \textbf{D-GCN} and \textbf{D-RGCN} are variants of our model that replace heterogeneous graph layers with  homogeneous graph layers, including GAT \cite{gat}, GCN \cite{gcn} and RGCN \cite{rgcn}. Note that D-GAT also use message fusion module to update representations for utterance nodes.

\subsection{Automatic Evaluation}

\begin{table}[t]
\centering
    \begin{tabular}{c|l|c|c|c|c|ccc}
        \toprule
        \textbf{Type} & \textbf{Model}& \textbf{Know.} & \textbf{Heter.} & \textbf{Utter.} & \textbf{RL}& \textbf{R-1} & \textbf{R-2} & \textbf{R-L}  \\
        \midrule
        \multirow{3}{*}{Extractive} & LONGEST-3 &\ding{55} &\ding{55} &\ding{55} &\ding{55} &32.46 &10.27 &29.92 \\
        & TextRank & \ding{55} &\ding{55} &\ding{55}  &\ding{55} &29.27 &8.02 &28.78 \\
        & SummaRunner &\ding{55} &\ding{55} &\ding{55} &\ding{55} &33.76 &10.28  &28.69 \\
        \midrule  		 
        \multirow{3}{*}{Abstractive} &Transformer &\ding{55} &\ding{55} &\ding{55} &\ding{55} &36.62 &11.18  &33.06 \\
        &PGN &\ding{55} &\ding{55} &\ding{55} &\ding{55} &40.08 &15.28 &36.63 \\
        &HRED &\ding{55} &\ding{55} &\ding{55} &\ding{55} &40.39 &16.13 &37.65 \\
        \midrule
        \multirow{2}{*}{\makecell{Pipeline}} &Abs RL &\ding{55} &\ding{55} &\ding{51} &\ding{51} &40.96 &17.18 &39.05 \\
        &AbsRL Enhance &\ding{55} &\ding{55} &\ding{51} &\ding{51} &41.95 &18.06 &39.23 \\
        \midrule  		 
        \multirow{4}{*}{Ours} &D-GCN &\ding{51} &\ding{55} &\ding{55} &\ding{55} &41.33 &16.98 &38.70 \\
        &D-GAT &\ding{51} &\ding{55}  &\ding{55} &\ding{55} &41.08 &16.89 &38.61 \\
        &D-RGCN &\ding{51} &\ding{55} &\ding{55} &\ding{55} &41.36 &17.07 &38.93 \\
        \cmidrule{2-9}  
        &D-HGN &\ding{51} &\ding{51} &\ding{55} &\ding{55} &\textbf{42.03} &\textbf{18.07} &\textbf{39.56} \\
        \bottomrule 
    \end{tabular}
\caption{Test set results on the SAMSum Dataset, where ``R-1'' is short for ``ROUGE-1'', ``R-2'' for ``ROUGE-2'', ``R-L'' for ``ROUGE-L''. ``Know.", ``Heter.", ``Utter." and ``RL" indicate whether knowledge, heterogeneity modeling, utterance-level extraction labels and reinforcement learning are used or not.} \label{tab:main_results}
\end{table}

Table \ref{tab:main_results} shows the results on SAMSum corpus. The D-HGN stands for our full model, which outperforms various baselines. Compared with HRED that uses no additional auxiliary information such as commensence knowledge or utterance-level extraction labels, D-RGCN that uses commensence knowledge can achieve 0.97\% improvement on ROUGE-1, 0.94\% on ROUGE-2, 1.28\% on ROUGE-L, which shows the effectiveness of knowledge integration. Compared with homogeneous networks like D-RGCN, D-HGN that based on heterogeneous graph networks can achieve 0.67\% improvement on ROUGE-1, 1.00\% on ROUGE-2, 0.63\% on ROUGE-L, which verifies the effectiveness of heterogeneity modeling.

\subsection{Ablation Study}
We conduct two types of ablation studies to verify the effectiveness of different types of nodes and two modules we propose. As shown in Table \ref{tab:ab}(a), without knowledge integration(w/o knowledge), the model suffers the performance drop, which shows incorporating knowledge can help our model better modeling the dialogue context. For speaker nodes, directly remove them in the graph will lead to no speaker in the final summary. Instead, we append the speakers in front of utterances(w/o speaker). The results show that modeling speakers as heterogeneous data will do good the final summary generation process. As shown in Table \ref{tab:ab}(b), we remove the message fusion module(w/o message fusion), the results show that it is worth to design specific message fusion method according to different types of nodes. Besides, without taking position information into account(w/o node embedding), our model will lose some performance.

\begin{table}[!htb]
    \begin{subtable}{.5\linewidth}
      \centering
        \begin{tabular}{lccc}
        \toprule
        \textbf{Model} & \textbf{R-1} & \textbf{R-2} & \textbf{R-L}  \\
        \midrule  		 
        D-HGN &\textbf{42.03} &\textbf{18.07} &\textbf{39.56}  \\
          \ \  w/o \textit{knowledge} &41.52 &17.38 &38.76 \\	
          \ \  w/o \textit{speaker} &41.06 &17.17 &38.92\\
        \bottomrule 
        \end{tabular}
        \caption{Ablation Study for Different Types of Nodes}
    \end{subtable}%
    \begin{subtable}{.5\linewidth}
      \centering
        \begin{tabular}{lccc}
        \toprule
        \textbf{Model} & \textbf{R-1} & \textbf{R-2} & \textbf{R-L}  \\
        \midrule  		 
        D-HGN &\textbf{42.03} &\textbf{18.07} &\textbf{39.56}  \\
          \ \  w/o \textit{message fusion} &41.29 &17.09 &38.74 \\
          \ \  w/o \textit{node embedding} &41.99 &17.85 &38.89 \\	
        \bottomrule 
        \end{tabular}
        \caption{Ablation Study for Two Modules} 
    \end{subtable} 
    \caption{Ablation Study.} \label{tab:ab}
\end{table}

\subsection{Human Evaluation}
We conduct human evaluation to verify the quality of the generated summaries, including abstractiveness (contains higher-level conceptual words), informativeness (covers adequate information) and correctness (associates right names with actions). We hired five graduates to perform human evaluation. For each metric, the score ranges from 1 (worst) to 5 (best). The results are shown in Table \ref{tab:he}.

\begin{table}[htb]
\centering
        \begin{tabular}{lcccc}
            \toprule
            \textbf{Model} & \textbf{Abstractiveness} & \textbf{Informativeness} & \textbf{Correctness}  \\
            \midrule 
            PGN    &2.70 &2.68 &2.49  	 	  \\
            AbsRL Enhance    &2.94 &3.23  &2.43   \\
            \midrule  
            D-HGN    &\textbf{3.26} &\textbf{3.25}  &\textbf{2.92}  	 	  \\
            \ \  w/o \textit{knowledge} &3.09 &3.16 &2.80  	  \\
            \ \  w/o \textit{speaker} &3.23 &3.21 &2.60  	 	  \\
            \bottomrule 
        \end{tabular}
\caption{Human evaluation results.}
\label{tab:he}
\end{table}

Our model achieves higher scores. Compared with D-HGN, D-HGN(w/o knowledge) gets a lower score in abstractiveness, which indicates knowledge incorporation can help our model express deeper meanings. D-HGN(w/o speaker) performs worse than D-HGN in correctness, which shows effectiveness of heterogeneity modeling by viewing speakers as heterogeneous data. AbsRL Enhance performs worst in correctness, which may due to the utterances extraction  will break the coherence of dialogue contexts. 

\subsection{Zero-shot Setting}
To verify whether knowledge can help our model better generalize to the new domain, we directly test models on the ADSC Corpus. The results are shown in Table \ref{tab:adsc_results}. The homogeneous model D-GAT that uses knowledge can get better results than other baselines. The D-HGN gets the best score. We contribute this to the fact that knowledge can help our models better understand the dialogue in the new domain.

\begin{table}[htb]
\centering
    \begin{tabular}{lccc}
        \toprule
        \textbf{Model} & \textbf{ROUGE-1} & \textbf{ROUGE-2} & \textbf{ROUGE-L}  \\
        \midrule
        PGN & 28.69 &4.77 &	22.39  \\
        AbsRL Enhance  & 30.00 	&4.87 	&22.27 \\
        \midrule  
        D-GAT &32.90 &5.46 &22.47 \\
        D-HGN &\textbf{33.55} &\textbf{5.68} &\textbf{22.75} \\
        \bottomrule 
    \end{tabular}
\caption{ROUGE $F_1$ results on the Argumentative Dialogue Summary Corpus.} \label{tab:adsc_results}
\end{table}

\subsection{Visualization}
\begin{figure}[htb]
	\centering
	\includegraphics[scale=0.24]{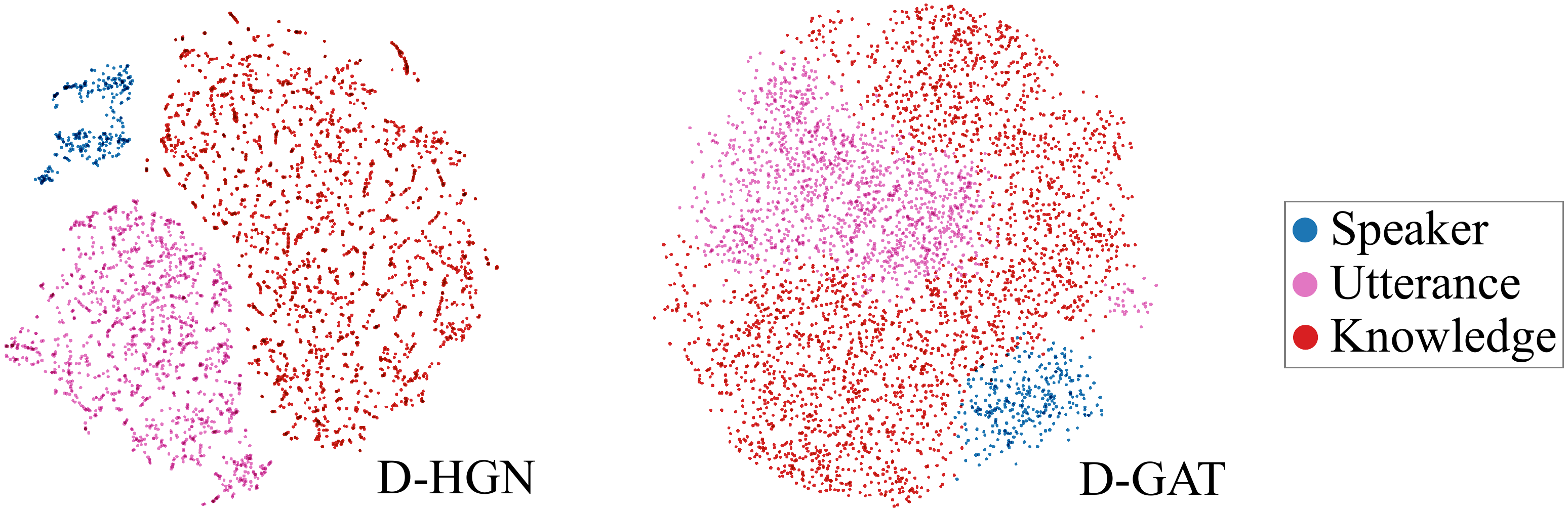}
	\caption{Visualization of node representations generated by the last graph layer of D-HGN and D-GAT.}
	\label{fig:emb_vis}
\end{figure}

To examine whether our D-HGN can learn easily distinguishable representations, we extract node representations from the last graph layer for the SAMSum test set. We apply t-SNE \cite{tsne} to these vectors. The results are shown in Figure \ref{fig:emb_vis}. We find that our model can generate more discrete and easily distinguishable representations. Besides, D-GAT also tends to separate representations of different types of nodes, which indicates explicitly heterogeneity modeling is a more reasonable approach.
 
\subsection{Case Study}
\begin{figure}[htb]
	\centering
	\includegraphics[scale=0.60]{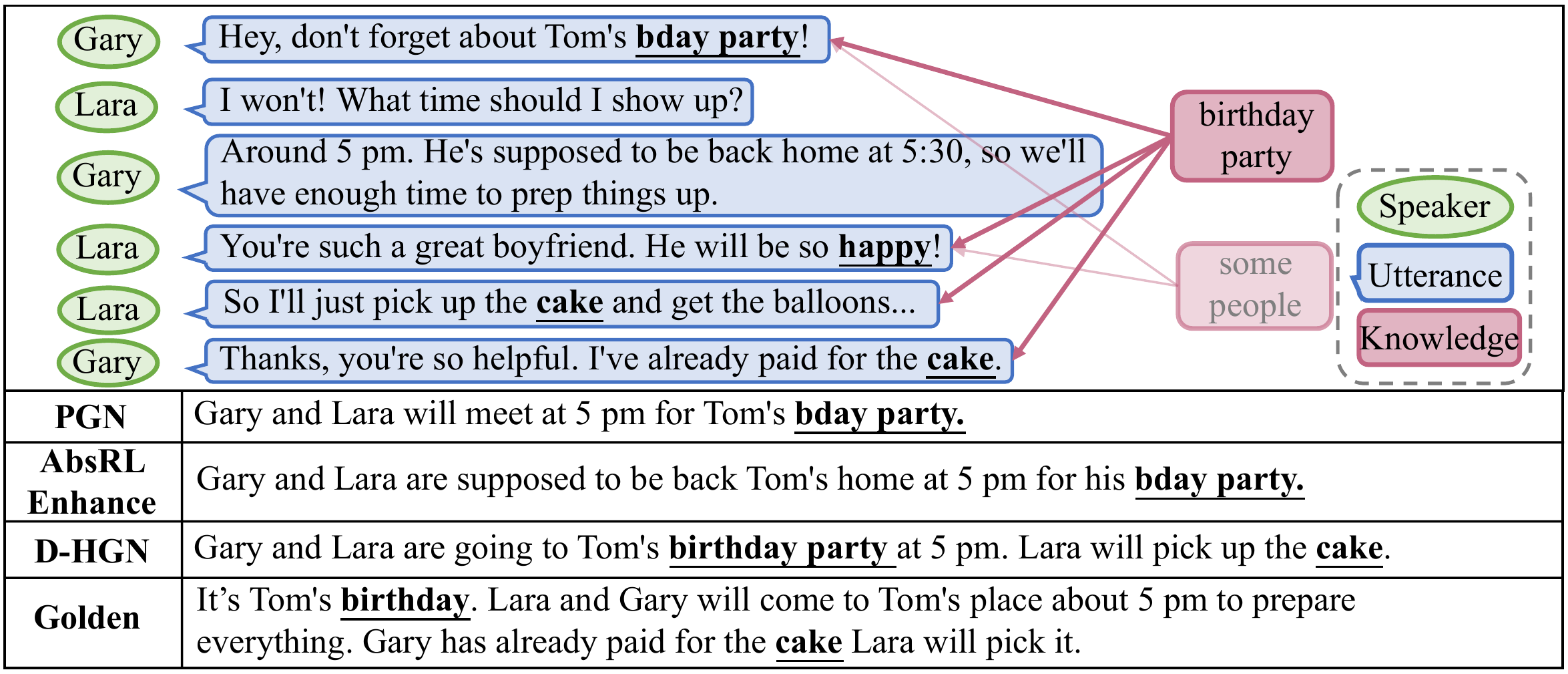}
	\caption{Example summaries generated by different models for one dialogue.}
	\label{fig:case}
\end{figure}

Figure \ref{fig:case} shows summaries generated by different models and the visualization of knowledge-to-utterance attention weights learned by our D-HGN model, the darker the color, the higher the weights. Our model incorporates two knowledge nodes, one is {\em birthday party} according to ``bday party", ``happy" and ``cake", the other one is {\em some people} according to ``Tom" and ``boyfriend". We can see that our D-HGN model pays more attention to {\em birthday party} rather than {\em some people}. On the one hand, incorporating {\em birthday party} helps our model generate a more formal summary (using birthday rather than bday). On the other hand, {\em birthday party} connects non-adjacent utterances around the birthday topic, which helps our model generate a more informative and detailed summary (including cake).

\section{Related Work}
Previous works used feature engineering \cite{xie2008evaluating}, template-based \cite{oya2014template} and graph-based \cite{bui2009extracting} methods for extractive dialogue summarization. Although extractive methods are widely used, the results tend to be incoherent and poorly readable. Therefore, current works mainly focus on abstractive methods, which can produce more readable and ﬂuency summaries. They tend to incorporate additional auxiliary information to help better modeling the dialogue. \newcite{sentgate} incorporated dialogue acts to model the interactive status of the meeting. \newcite{kdd19} tackled the problem of customer service summarization, which first produced a sequence of pre-defined keywords then generated the summary. \newcite{topic-dialogue} generated summaries for nurse-patient conversation by incorporating topic information. \newcite{crf-discourse} first removed useless utterances by utilizing discourse labels and then generated summaries. \newcite{acl19li} combined vision and textual features in a unified hierarchical attention framework to generate meeting summaries. \newcite{hmnet} employed a hierarchical transformer framework and incorporated part-of-speech and entity information for meeting summarization. In this paper, we facilitate dialogue summarization task by incorporating commonsense knowledge and further model utterances, commonsense knowledge and speakers as heterogeneous data.

\section{Conclusion}
In this paper, we improve abstractive dialogue summarization by incorporating commonsense knowledge. We first construct a heterogeneous dialogue graph by introducing knowledge from a large-scale commonsense knowledge base. Then we present a Dialogue Heterogeneous Graph Network (D-HGN) for this task by viewing utterances, knowledge and speakers in the graph as heterogeneous nodes. We additionally design two modules named message fusion and node embedding to facilitate information flow. Experiments on the SAMSum dataset show the effectiveness of our model that can outperform various methods. Zero-shot setting experiments on the Argumentative Dialogue Summary Corpus show that our model can better generalized to the new domain.

\bibliographystyle{coling}
\bibliography{coling2020}

\end{document}